\documentclass[final,preprint]{elsarticle}
\usepackage{lineno}

\modulolinenumbers[5]

\journal{Applied Soft Computing}

\usepackage{amsmath}
\usepackage{amssymb}
\usepackage{amsfonts}

\usepackage{booktabs}
\usepackage{multirow}
\usepackage{subfig}
\usepackage{floatflt}
\usepackage{wrapfig}
\usepackage[linesnumbered, ruled,vlined]{algorithm2e}
\usepackage{url}
\usepackage{xcolor}
\usepackage{adjustbox}
\usepackage{tablefootnote}
\usepackage{lscape}
\usepackage{pdflscape}
\usepackage{moresize}

\def\vector#1{\mbox{\boldmath $#1$}}

\begin{document}

\begin{frontmatter}

\title{An Easy-to-use Real-world Multi-objective Optimization Problem Suite}


\author{Ryoji Tanabe}
\ead{rt.ryoji.tanabe@gmail.com}

\author{Hisao Ishibuchi\corref{cor1}}
\ead{hisao@sustech.edu.cn}
\cortext[cor1]{Corresponding author}

\address{Shenzhen Key Laboratory of Computational Intelligence, University Key Laboratory of Evolving Intelligent Systems of Guangdong Province, Department of Computer Science and Engineering, Southern University of Science and Technology, Shenzhen 518055, China}







\begin{abstract}

Although synthetic test problems are widely used for the performance assessment of evolutionary multi-objective optimization algorithms, they are likely to include unrealistic properties which may lead to overestimation/underestimation.
To address this issue, we present a multi-objective optimization problem suite consisting of 16 bound-constrained real-world problems.
The problem suite includes various problems in terms of the number of objectives, the shape of the Pareto front, and the type of design variables.
4 out of the 16 problems are multi-objective mixed-integer optimization problems.
We provide Java, C, and Matlab source codes of the 16 problems so that they are available in an off-the-shelf manner.
We examine an approximated Pareto front of each test problem.
We also analyze the performance of six representative evolutionary multi-objective optimization algorithms on the 16 problems.
In addition to the 16 problems, we present 8 constrained multi-objective real-world problems.

\end{abstract}


\end{frontmatter}

\nolinenumbers

\section{Introduction}
\label{sec:introduction}





A bound-constrained multi-objective optimization  problem (MOP) is to find a solution $\vector{x} \in \mathbb{S} \subseteq \mathbb{R}^D$ that minimizes an objective function vector $\vector{f}: \mathbb{S} \rightarrow \mathbb{R}^M$.
Here, $\mathbb{S}$ is the $D$-dimensional solution space, and $\mathbb{R}^M$ is the $M$-dimensional objective space.
In general, the goal of MOPs is to find a set of non-dominated solutions that approximates the Pareto front in the objective space.


Evolutionary multi-objective optimization algorithms (EMOAs) are effective approaches for MOPs \cite{Deb01}.
It is expected that an EMOA is able to find a set of non-dominated solutions with acceptable quality for a decision maker in a single run.
A number of EMOAs have been proposed, including NSGA-II \cite{DebAPM02}, SPEA2 \cite{ZitzlerLT01}, IBEA \cite{ZitzlerK04}, SMS-EMOA \cite{BeumeNE07}, MOEA/D \cite{ZhangL07}, and NSGA-III \cite{DebJ14}.
In general, it is difficult to theoretically evaluate the performance of EMOAs mainly due to their stochastic nature.
Thus, the performance assessment of EMOAs is empirically conducted on benchmark problems in most cases.

Test problems play a crucial role in the empirical performance evaluation.
Synthetic test problems are usually used in the EMO community.
Representative examples include the ZDT \cite{ZitzlerDT00}, DTLZ \cite{DebTLZ05}, WFG \cite{HubandHBW06}, and LZ \cite{LiZ09} problem suites.
The main advantages of synthetic test problems are threefold.
First, they are easy to use.
Since synthetic test problems are expressed by simple equations, they can be easily implemented in any programming language.
The calculation of each objective function is fast.
Second, the properties of synthetic test problems are (almost) clear.
Their Pareto fronts are usually known.
Thus, some general knowledge can be obtained from experimental results on them.
For example, an EMOA that works well on the DTLZ1 problem \cite{DebTLZ05} is likely to have a good performance on separable multi-modal problems whose Pareto fronts are linear and triangular.
In contrast, the properties of real-world problems are generally unclear.
Thus, results on most real-world problems cannot be generalized.
Third, some synthetic problems are scalable to the number of objectives $M$ and design variables $D$.
The scalability of EMOAs to $M$ and $D$ can be evaluated on such scalable problems.





\begin{figure}[!b]
  \newcommand{\widthvar}{0.25}
\centering
\subfloat[Triangular-linear]{\includegraphics[width=\widthvar\textwidth]{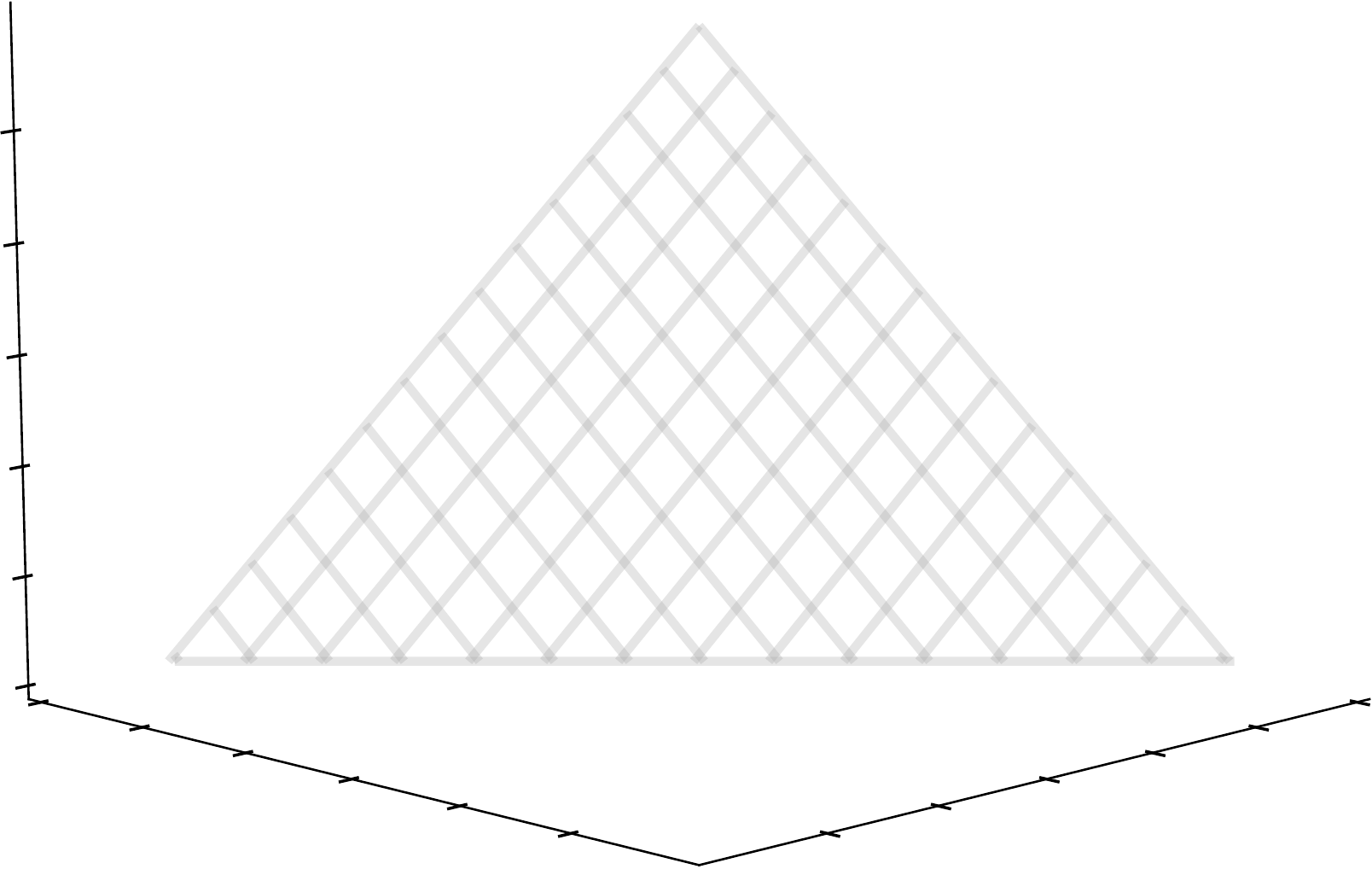}}
\subfloat[Triangular-concave]{\includegraphics[width=\widthvar\textwidth]{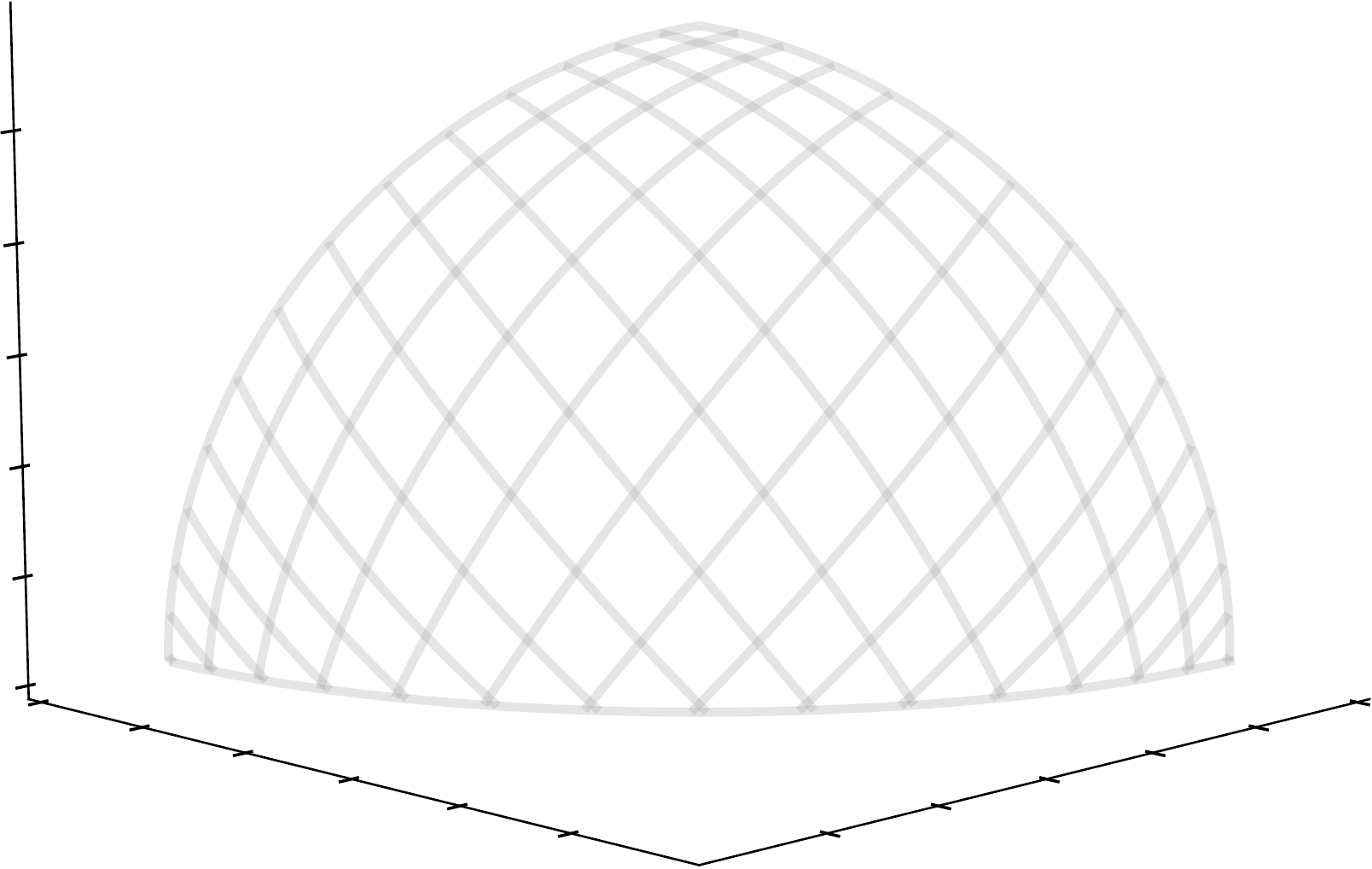}}
\subfloat[Triangular-convex]{\includegraphics[width=\widthvar\textwidth]{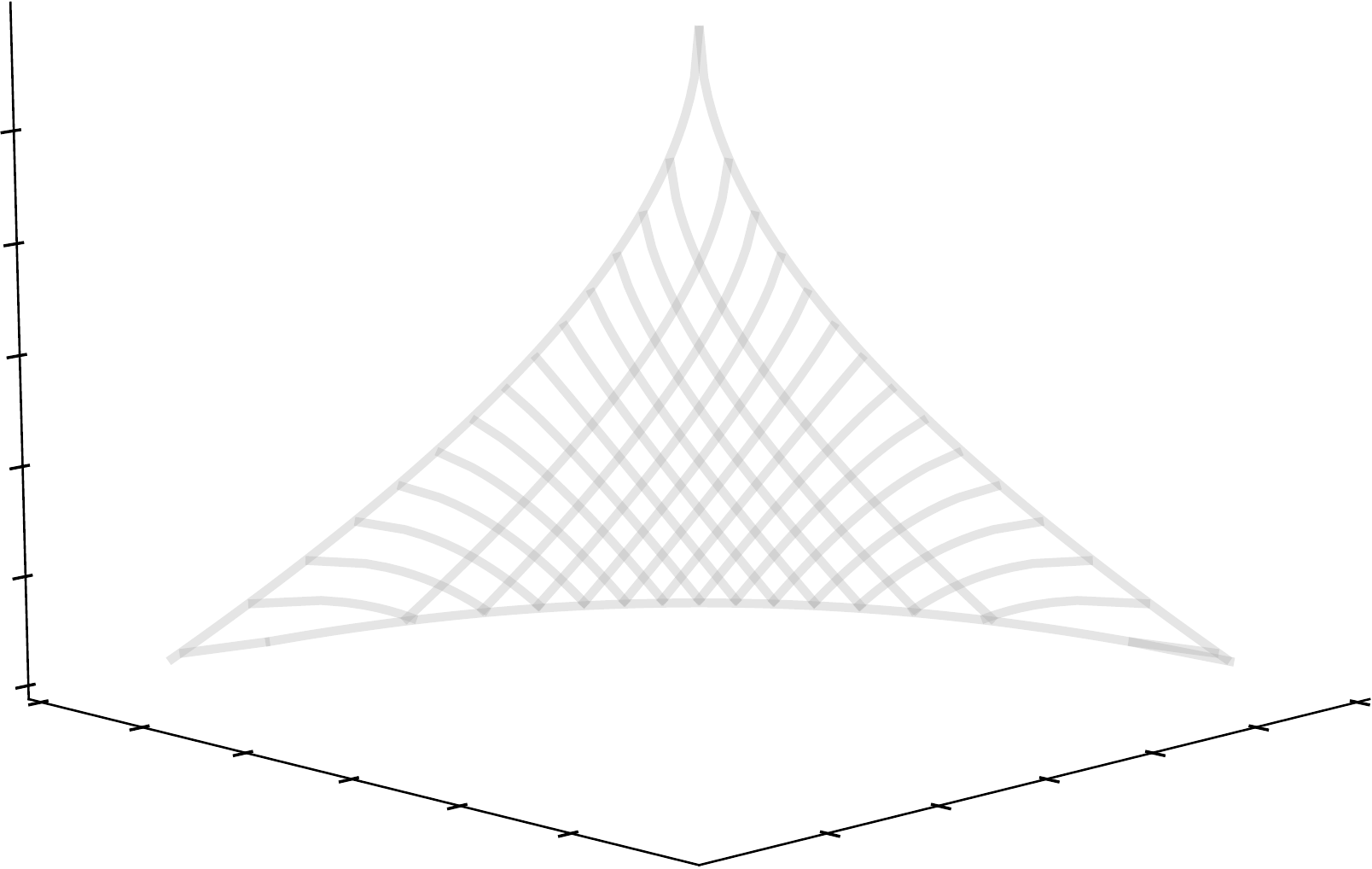}}
\caption{Triangular shape Pareto fronts. The $x$, $y$, and $z$ axis represent $f_1$, $f_2$, and $f_3$, respectively.}
\label{fig:pfshape}
\end{figure}

Despite these advantages, most synthetic test problems have difficulties.
They have unusual properties which are unlikely to appear in real-world applications \cite{IshibuchiSMN16,ZapotecasCAT18}.
The performance of an EMOA exploiting those unrealistic properties could be overrated on such problems.
For example, some decomposition-based EMOAs (e.g., MOEA/D \cite{ZhangL07} and NSGA-III \cite{DebJ14}) work well on the DTLZ and WFG problems even with many objectives \cite{IshibuchiSMN16}.
This is because most DTLZ and WFG problems have triangular shape Pareto fronts.
The term ``triangular'' means that the shape of the Pareto front looks like a distorted/bended triangle in a three-dimensional objective space \cite{IshibuchiSMN16}.
Many existing scalable test problems have such triangular shaped Pareto fronts.
Figures \ref{fig:pfshape}(a)--(c) show linear, concave, and convex triangular shape Pareto fronts under three objectives, respectively.
The Pareto fronts shown in Figures \ref{fig:pfshape}(a)--(c) are identical to those of DTLZ1, DTLZ2 (DTLZ3, DTLZ4, WFG4--9), and convex DTLZ2 \cite{DebJ14}.
Any objective vector $\vector{f}(\vector{x})$ in the Pareto front satisfies the following condition in the normalized objective space $[0,1]^M$ \cite{IshibuchiHS19}: $\sum^{M}_{i=1}\bigl(f_i(\vector{x})\bigr)^p=1$, where $0 \leq \bigl(f_i(\vector{x})\bigr) \leq 1$ for $i \in \{1, ..., M\}$, $p=1$ for the linear Pareto front, $1<p$ for the concave Pareto front, and $0<p<1$ for the convex Pareto front.
The shape of the distribution of weight vectors in decomposition-based EMOAs is highly consistent with the shape of the triangular Pareto front \cite{IshibuchiSMN16}.
For this reason, some decomposition-based EMOAs perform well on most DTLZ and WFG test problems even with many objectives.
In contrast, some decomposition-based EMOAs perform poorly on problems with non-triangular Pareto fronts (e.g., the inverted DTLZ problems \cite{JainD14}).
%
%
Test problems designed with the bottom-up approach (e.g., the DTLZ and WFG problems) include the distance and position variables.
Here, the distance variables control the distance between the objective vector and the Pareto front.
The position variables determine the position of the objective vector on the Pareto front.
Since the distance and position variables play different roles, it is relatively easy to improve the convergence and diversity of a set of solutions by individually changing the distance and position variables \cite{IshibuchiSMN16}.

Since synthetic test problems often exhibit such undesirable features, a new EMOA should be benchmarked on some real-world problems for more reliable evaluation.
However, there does not exist an easy-to-use real-world multi-objective optimization problem suite.
Although EMOAs have been applied to a number of real-world problems \cite{ObayashiSTH00,PonsichJC13,SinghRS13,ChengRFOJ17}, it is difficult to use real-world problems to evaluate the performance of EMOAs.

In most cases, a new method is generally benchmarked only on some synthetic test problems.
For example, the previous study \cite{ZhangL07} evaluated the performance of MOEA/D on ZDT and DTLZ.
In some cases, a new method is benchmarked on a few real-world problems in addition to synthetic test problems.
For example, the previous study \cite{DebAPM02} evaluated the performance of NSGA-II on the water resource planning problem  \cite{RayTS01}.
However, only a few real-world problems are generally used for benchmarking EMOAs in almost all previous studies.
Thus, the general performance of EMOAs on various real-world problems has not been examined in the literature.
On the one hand, as mentioned above, EMOAs have been applied to a number of real-world problems, including supersonic wing design problems \cite{ObayashiSTH00}, finance/economics application problems \cite{PonsichJC13}, oil production planning problems \cite{SinghRS13}, and electric vehicle control problems \cite{ChengRFOJ17}.
On the other hand, comparisons of multiple EMOAs were not performed in most application-oriented studies.
An application-oriented study usually reports experimental results of a single EMOA in order to demonstrate that a good approximation of the Pareto front of a given real-world problem can be obtained (i.e., a better approximation can be obtained by the EMO approach than other approaches).
Thus, the performance of multiple EMOAs on a given real-world problem was not investigated in such an application-oriented study.
%
The RE problem set makes the performance comparison on real-world problems easy.

Let us consider the performance comparison of EMOAs on supersonic wing design problems \cite{ObayashiSTH00}.
In the first place, researchers who have no knowledge of aerodynamics cannot tackle this real-world problem.
Numerical simulations of complex fluid flow require specific software tools, high-performance computers, and long computation time.
According to \cite{ObayashiSTH00}, it took about $100$ hours to evaluate $4\,480$ solutions on the supersonic wing design problems using 32 processing elements of an NEC SX-4 computer.
If six EMOAs are carried out on this problem for $30$ times, it takes about $2$ years in total.
While benchmarking EMOAs on real-world problems is needed, it is difficult due to those obstacles.


\begin{table}[b]
\begin{center}
  \caption{\small Advantages and disadvantages of each problem type.}
{\footnotesize
  \label{tab:problems_pros_cons}
\scalebox{1}[1]{ 
\begin{tabular}{lccccc}
\midrule
Problems & Non-artificiality & Easiness & Clearness & Scalability\\ 
\toprule
Synthetic problems & & $\checkmark$ & $\checkmark$ & $\checkmark$\\\midrule
RE problems  & $\checkmark$ & $\checkmark$ & & \\\midrule
\end{tabular}
}
}
\end{center}
\end{table}




%



To fill this gap between the necessity and the availability of real-world problems, we present 16 bound-constrained real-world multi-objective optimization problems where definitions are given by simple equations or surrogate models.
We refer to the set of the 16 problems as the \underline{RE}al world problem suite RE.
The RE problem suite provides a wide variety of computationally cheap problems regarding the number of objectives, the shape of the Pareto front, and the type of design variables.
The 16 RE problems do not perform a computationally expensive simulation.
As shown in Section S.1 in the supplementary file of this paper, the objective functions in the 16 RE problems are represented by relatively simple equations.
For this reason, the objective functions in the 16 RE problems are computationally cheap.
Table \ref{tab:problems_pros_cons} shows a summary of the advantages and disadvantages of synthetic problems and the RE problems.
In contrast to synthetic problems, we believe that the RE problems do not have the above-mentioned unrealistic properties (e.g., the triangular Pareto front).
We do not claim that real-world problems (including the RE problems) must always be better than synthetic test problems in terms of the performance evaluation of evolutionary algorithms.
As mentioned above, most synthetic test functions have at least three advantages.
Although the RE problems are as easy to use as synthetic test functions, they have two disadvantages compared to synthetic test functions.
Similar to other real-world problems, the properties of the RE problems are unclear.
All the RE problems are not scalable with respect to the number of objectives $M$ and design variables $D$.
However, we try to address the two disadvantages of the RE problems in this paper.
We analyze the shape of the Pareto front of each RE problem in order to clarify its property.
We have selected a wide variety of real-world problems with respect to $M$ and $D$ so that the scalability of EMOAs can be evaluated.

Our main contributions in this paper are as follows:

\begin{enumerate}
\item We present the 16 RE problems. Despite their importance, most of the original 16 RE problems have not received much attention in the EMO community.
Four RE problems are multi-objective mixed-integer optimization problems. As pointed out in \cite{TusarBH19}, there are only a few (multi-objective) mixed-integer optimization problems for benchmarking evolutionary algorithms. For this reason, we believe that the four mixed-integer RE problems are helpful in investigating the performance of evolutionary algorithms for multi-objective mixed-integer optimization.
\item We present details of the 16 RE problems in the supplementary file of this paper. The supplementary file is also available at the supplementary website (\url{https://github.com/ryojitanabe/reproblems}). Since this paper is self-contained, readers do not need to refer to each original article.
\item We uploaded Java, C, and Matlab source codes of the RE problems to the supplementary website. Researchers can easily examine the performance of their EMOAs on the RE problems in an off-the-shelf manner.
\item We analyze the performance of six representative EMOAs on the RE problems.
\item In addition to the 16 RE problems, we present 8 \underline{C}onstrained multi-objective \underline{RE}al-world problems (CRE).
\end{enumerate}

The rest of this paper is organized as follows.
Section \ref{sec:probdefs} presents the 16 RE problems.
Section \ref{sec:experimental_settings} describes experimental settings for the comparisons of EMOAs on the RE problem set. 
Section \ref{sec:experimental_results} reports experimental results.
Section \ref{sec:conclusion} concludes this paper. 




\section{RE problem suite}
\label{sec:probdefs}



This section describes the RE problem suite.
Table \ref {tab:testproblems} shows the properties of the 16 RE problems.
Section S.1 in the supplementary file provides details of the RE problems, including their definitions.
In the first column of Table \ref{tab:testproblems}, we use the notation RE''$M$''-''$D$''-''$Z$'' in this paper, where $M$ is the number of objectives, $D$ is the number of decision variables, and $Z$ is the problem ID.
For example, $M$ and $D$ of RE3-4-2 are 3 and 4, respectively.
The problem ID prevents that different RE problems share the same name (e.g., RE3-4-2 and RE3-4-3).
%
If a new problem is proposed, it can be sequentially added to the RE problem set (e.g., a new three-objective and five-variable problem could be denoted as the RE3-5-8 problem).
The current RE problem suite consists of problems with  $M \in \{2, 3, 4, 6, 9\}$.



\subsection{On four surrogate-based RE problems}
\label{sec:surrogate-problems}

Four out of the 16 RE problems (RE3-5-4, RE3-4-7, RE4-7-1, and RE9-7-1) include parameters obtained by the response surface method using data sampled from simulations.
Thus, these four surrogate-based RE problems are not exactly the same as their corresponding original problems.
Although the four RE problems are based on data taken from the real-world applications, they may be viewed as ``artificial'' problems.
This fact should be kept in mind.




\subsection{Reformulation of constrained single- and multi-objective real-world problems}
\label{sec:reformulation}

The original car side-impact problem is a constrained single-objective problem with eight constraint functions \cite{GuYTMFL01}.
A nine-objective version of it is designed in \cite{DebJ14} by converting the constraint functions into eight new objective functions.
Inspired by \cite{DebJ14}, we reformulated constrained single- and multi-objective real-world problems as bound-constrained multi-objective problems in a similar manner (except for RE2-4-1, RE3-5-4, and R3-4-7). 
However, our preliminary results indicated that the constraint functions of most original problems in Table \ref {tab:testproblems} can be simultaneously minimized.
Thus, we formulated the sum of constraint violation values as an additional objective function for all RE problems except for RE9-7-1.

For example, RE-2-3-2 is the reinforced concrete beam design problem \cite{AmirH89}.
The first objective $f_1$ of the RE2-3-2 problem is to minimize the total cost of concrete and reinforcing steel of the beam:
\begin{align}
f_1 (\vector{x}) =  29.4 x_1 + 0.6 x_2 x_3,
\end{align}
where $x_2 \in [0, 20]$ and $x_3 \in [0, 40]$.
In RE-2-3-2, $x_1$ has a pre-defined discrete value from 0.2 to 15 (for details, see Section S.1.2 in the supplementary file).
The three variables $x_1, x_2$, and $x_3$ represent the area of the reinforcement, the width of the beam, and the depth of the beam, respectively.
The second objective $f_2$ of RE2-3-2 is the sum of the two constraint violations:
\begin{align}
f_2 (\vector{x}) &=  \sum^2_{i=1} \max\{g_i(\vector{x}), 0\},\\
g_1 (\vector{x}) &=  x_1 x_3 - 7.735 \frac{x^2_1}{x_2} - 180 \geq 0,\\
g_2 (\vector{x}) &=  4 - \frac{x_3}{x_2} \geq 0.
\end{align}
Due to space constraint, we explain details of the other 15 RE problems in the supplementary file.      
Interested readers can refer to the supplementary file.

\begin{table}[t]
\centering
\caption{\small Properties of the 16 test problems (RE2-4-1, ..., RE9-7-1). $M$ and $D$ denote the number of objectives  and the number of decision variables, respectively. The type of design variables, the shape of the Pareto front, and other information are also described.}
{\scriptsize
  \label{tab:testproblems}
\scalebox{0.93}[1]{ 
\begin{tabular}{llcccccc}
\toprule
Name & Original name & $M$ & $D$ & Variables & Pareto front\\
  \midrule
RE2-4-1 & Four bar truss design \cite{ChengL99} & 2 & 4 & Continuous & Convex \\
RE2-3-2 & Reinforced concrete beam design \cite{AmirH89} & 2 & 3 & Mixed & Mixed \\
RE2-4-3 & Pressure vessel design \cite{KannanK94} & 2 & 4 & Mixed & Mixed, Disconnected \\
RE2-2-4 & Hatch cover design  \cite{AmirH89} & 2 & 2 & Continuous & Convex \\
RE2-3-5 & Coil compression spring design  \cite{LampinenZ99} & 2 & 3 & Mixed & Mixed, Disconnected \\
RE3-3-1 & Two bar truss design  \cite{CoelloP05} & 3 & 3 & Continuous & Unknown \\
RE3-4-2 & Welded beam design \cite{RayL02} & 3 & 4 & Continuous & Unknown \\
RE3-4-3 & Disc brake design  \cite{RayL02} & 3 & 4 & Continuous & Unknown \\
RE3-5-4 & Vehicle crashworthiness design  \cite{LiaoLYZL08} & 3 & 5 & Continuous & Unknown \\
RE3-7-5 & Speed reducer design  \cite{FarhangMehrA02} & 3 & 7 & Mixed & Unknown \\
RE3-4-6 & Gear train design  \cite{DebS05a} & 3 & 4 & Integer & Concave, Disconnected \\
RE3-4-7 & Rocket injector design \cite{VaidyanathanTPS03} & 3 & 4 & Continuous & Unknown \\
RE4-7-1 & Car side impact design \cite{JainD14} & 4 & 7 & Continuous & Unknown \\
RE4-6-2 & Conceptual marine design  \cite{ParsonsS04} & 4 & 6 & Continuous & Unknown \\
RE6-3-1 & Water resource planning  \cite{RayTS01} & 6 & 3 & Continuous & Unknown \\
RE9-7-1 & Car cab design \cite{DebJ14} & 9 & 7 & Continuous & Unknown\\
\toprule
\end{tabular}
}
}
\end{table}


\subsection{CRE problem suite}
\label{sec:cre}

In addition to the 16 RE problems, we provide 8 constrained multi-objective real-world problems.
We denote the 8 constrained problems as constrained RE (CRE) problems.
Table \ref{tab:testproblems_c} shows the properties of the eight CRE problems (CRE2-4-1, ..., CRE5-3-1).
Each CRE problem is the original version of the corresponding RE problem (see Table \ref{tab:testproblems_c}).
Recall that an additional objective function for some RE problems is the sum of constraint violation values.
For the definition of each CRE problem, see the definition of the corresponding RE problem in Section S.1 in the supplementary file.

Studies on unconstrained multi-objective optimization problems can be the basis of studies on more challenging problems, including constrained multi-objective optimization problems \cite{Mezura-MontesC11}, dynamic multi-objective optimization problems \cite{NguyenYB12}, and multi-modal multi-objective optimization problems \cite{TanabeI19emmo}.
For this reason, as the first step toward designing a benchmark set of easy-to-use real-world problems, we focus only on unconstrained multi-objective optimization problems in this paper.
Although this paper does not analyze the eight CRE problems for this reason, we believe that the CRE problems could also be good benchmark problems for constrained multi-objective optimizers.


\subsection{Multi-objective mixed-integer RE problems}
\label{sec:mip_re}

In Table \ref {tab:testproblems}, some (or all) design variables of the five RE problems take integer or discrete values.
RE-2-3-2, RE2-4-3, RE2-3-5, and RE3-7-5 are multi-objective mixed-integer optimization problems, and RE3-4-6 is a multi-objective integer optimization problem.
%
%
For example, as explained in Section \ref{sec:reformulation}, the first variable $x_1$ in RE2-3-2 is the area of the reinforcement, which should be a pre-defined discrete value from $0.2$ to $15$.
It should be noted that the 16 RE problems in Table \ref{tab:testproblems} do not have categorical variables, whose ordering information cannot be exploited \cite{HutterHLS09}, such as a material of the designed product (e.g., copper, aluminum, and iron).
%
%
Although multi-objective mixed-integer optimization problems appear in real-world applications \cite{AlvesC07,BrownleeW15}, they have received less attention in the EMO community.
One reason could be the unavailability of the proper test problems.
The five RE problems would encourage researchers to develop new EMOAs for multi-objective mixed-integer optimization problems.

\subsection{Comparison to simulation-based real-world problems}
\label{sec:mip_re}

All the RE problems are derived from the literature published in the 1980s--2000s.
Also, the maximum number of decision variables in the RE problems is seven.
These reasons are due to the progress of computer technology.
Since some researchers could not use high-performance simulators in an easy manner twenty and thirty years ago, they needed to implement their real-world problems by human-understandable equations.
It was also difficult to formulate a problem with many decision variables without degrading the understandability.
Some RE problems are ``secondary products'' from these limitations.
In contrast, thanks to technological progress, recent researchers can easily perform complex simulations on their personal workstations.
As a result, almost all recently proposed real-world problems are based on computational simulations.
However, such simulation-based real-world problems cannot be added to the RE problem set since their problem formulations are not given by human-understandable equations.
These are the reasons why all the RE problems are derived from the old literature, and the maximum number of variables in the RE problems is seven.



\begin{table}[t]
  \centering
\caption{\small Properties of the eight constrained test problems (CRE2-3-1, ..., RE5-3-1).}
{\scriptsize
  \label{tab:testproblems_c}
\scalebox{0.93}[1]{ 
\begin{tabular}{llcccccc}
\toprule
Name & Original name & RE ver. & $M$ & $D$ & $N$ & Variables\\
  \midrule
CRE2-3-1 & Two bar truss design \cite{CoelloP05} & RE3-3-1 & 2 & 3 & 3 & Continuous \\
CRE2-4-2 & Welded beam design \cite{RayL02} & RE3-4-2 & 2 & 4 & 4 & Continuous \\
CRE2-4-3 & Disc brake design \cite{RayL02} & RE3-4-3 & 2 & 4 & 4 & Continuous\\
CRE2-7-4 & Speed reducer design \cite{FarhangMehrA02} & RE3-7-5 & 2 & 7 & 11 & Mixed \\
CRE2-4-5 & Gear train design \cite{DebS05a} & RE3-4-6 & 2 & 4 & 1 & Integer \\
CRE3-7-1 & Car side impact design \cite{JainD14} & RE4-7-1 & 3 & 7 & 10 & Continuous \\
CRE3-6-2 & Conceptual marine design \cite{ParsonsS04} & RE4-6-2 & 3 & 6 & 9 & Continuous \\
CRE5-3-1 & Water resource planning \cite{RayTS01} & RE6-3-1 & 5 & 3 & 7 & Continuous \\
\toprule
\end{tabular}
}
}
\end{table}

\begin{table}[b]
\begin{center}
  \caption{\small Five simulation-based real-world problems that are available online.}
{\footnotesize
  \label{tab:other_problems}
\scalebox{1}[1]{ 
\begin{tabular}{lccccc}
\toprule
Test problems & $M$ & $D$\\ 
\midrule
Radar waveform design problem \cite{Hughes06} & 9 & 4, ..., 12\\
Heat exchanger design problem \cite{DanielsRETF18} & 2 & Any\\
Car structure design problem \cite{KohiraKOT18} & 2 & 222\\
TopTrumps problem \cite{VolzNKT19} & 2 & 88, 128, 168, 208\\
MarioGAN problem \cite{VolzNKT19} & 2 & 10, 20, 30, 40 \\
\toprule
\end{tabular}
}
}
\end{center}
\end{table}

Table \ref{tab:other_problems} shows five simulation-based real-world problems that are available through the Internet. 
The five problems in Table \ref{tab:other_problems} were proposed recently.
The five problems have a larger number of decision variables than the RE problems.
However, unlike the RE problems, the five problems need to perform computational simulations to calculate the objective function values of a solution.
Thus, their problem formulations are not explicitly given by equations.
Also, some of them require specific software tools.
For example, the software for the radar waveform design problem \cite{Hughes06} can be executed only by Matlab.
For these reasons, we do not add the five problems into our RE problems.
It should be recalled that one of the advantages of the RE problems is easy-to-use.
This paper presents the RE problems so that researchers in the EMO community can use the 16 RE problems in an off-the-shelf manner.

\subsection{Approximated Pareto fronts of the RE problems}
\label{sec:re_pf}


Here, we discuss the shape of the approximated Pareto front of each RE problem.
Since the Pareto optimal solution set and the Pareto front are unknown in each RE test problem, we used an approximate Pareto solution set $\vector{A}^{\approx}$ for each problem.
$\vector{A}^{\approx}$ is available at the supplementary website.
We selected $500 \times M$ solutions in $\vector{A}^{\approx}$ from a large number of non-dominated solutions $\vector{A}$ found by various optimizers.
We confirmed that the size of $\vector{A}$ is more than $500 \times M$ for all 16 RE problems, except for RE3-4-6.
Due to the discrete property of RE3-4-6, the number of non-dominated solutions found in our experiments is limited to only 28.
Only for RE3-4-6, we set $\vector{A}$ to $\vector{A}^{\approx}$ as is.
For details of $\vector{A}$, see Section \ref{sec:performance_indicators}.
We used a distance-based method to select well-distributed non-dominated solutions in the objective space \cite{TanabeIO17}.
Section S.2 in the supplementary file explains the distance-based selection method.
%

\begin{figure}[t]
\centering
\subfloat[RE2-4-1]{\includegraphics[width=0.3\textwidth]{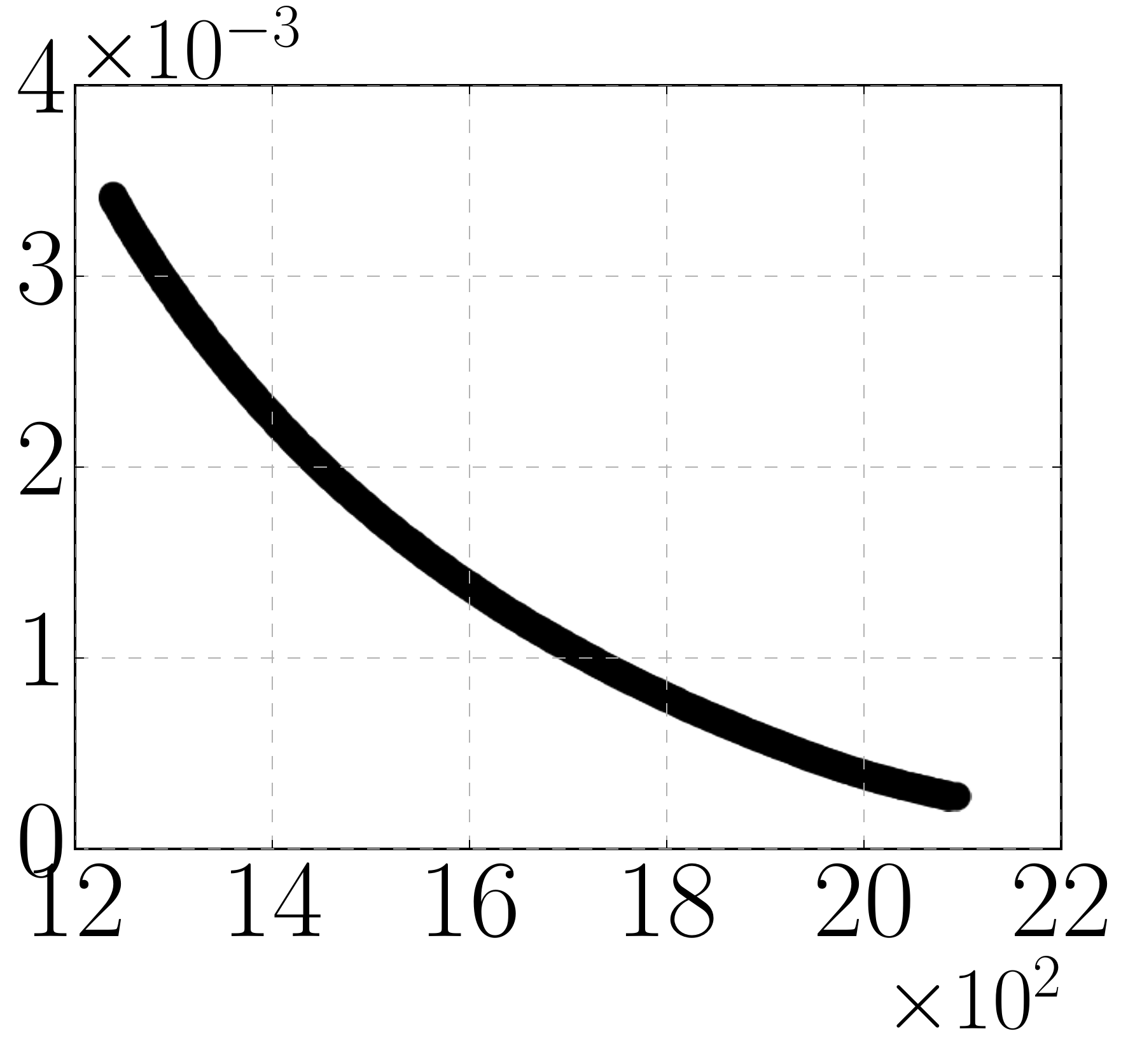}}
\subfloat[RE2-3-2]{\includegraphics[width=0.333\textwidth]{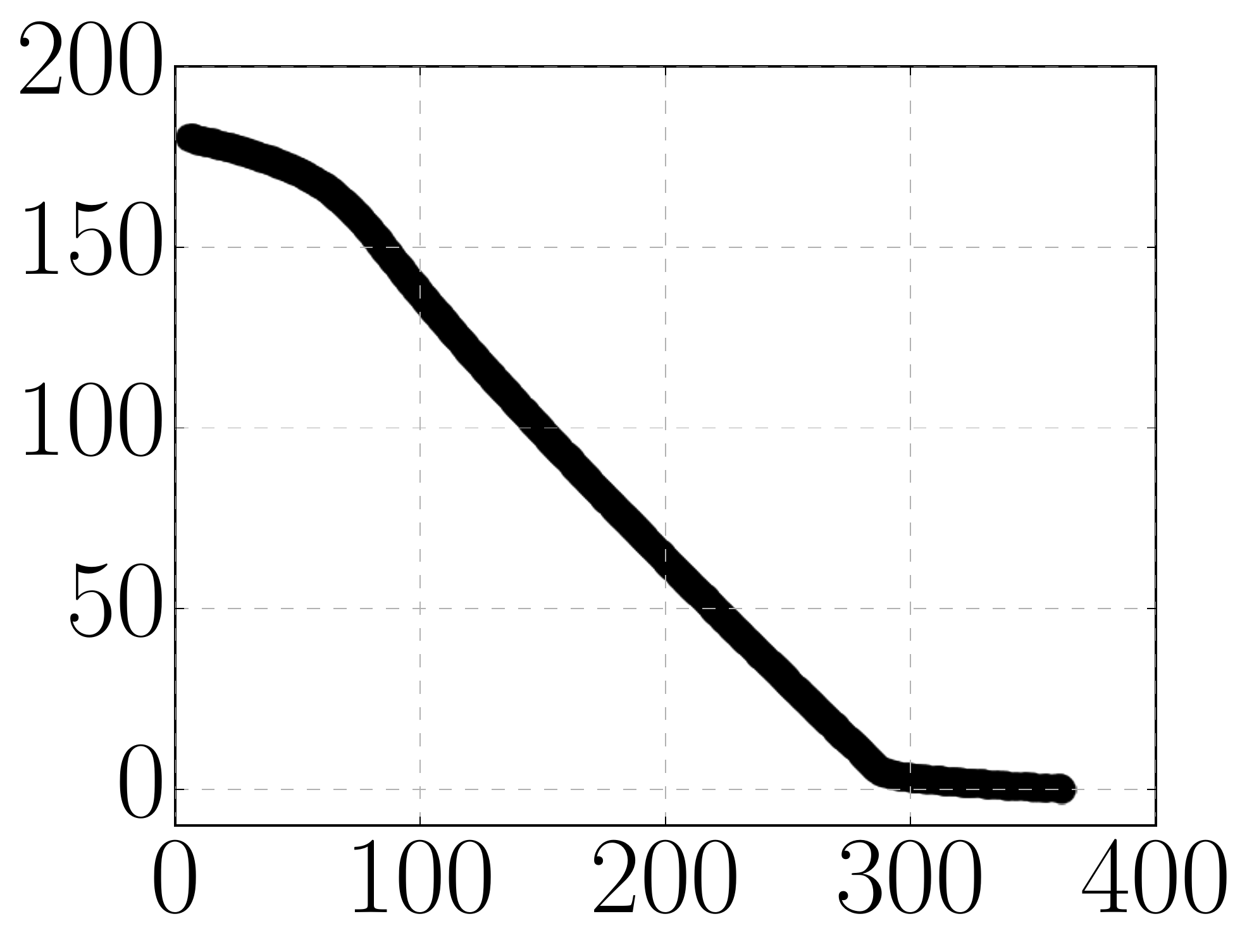}}
\subfloat[RE2-4-3]{\includegraphics[width=0.305\textwidth]{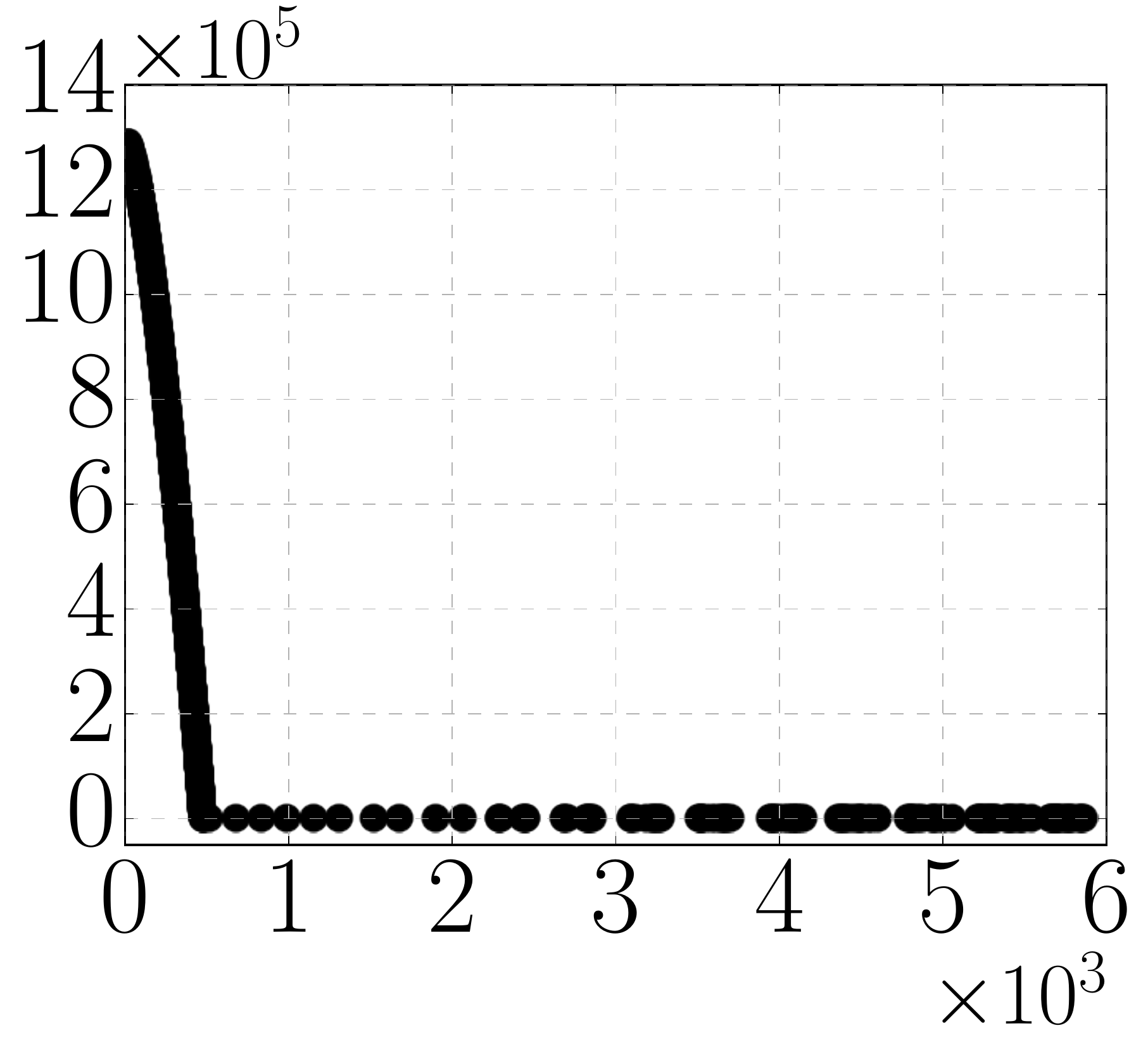}}
\\
\subfloat[RE2-2-4]{\includegraphics[width=0.33\textwidth]{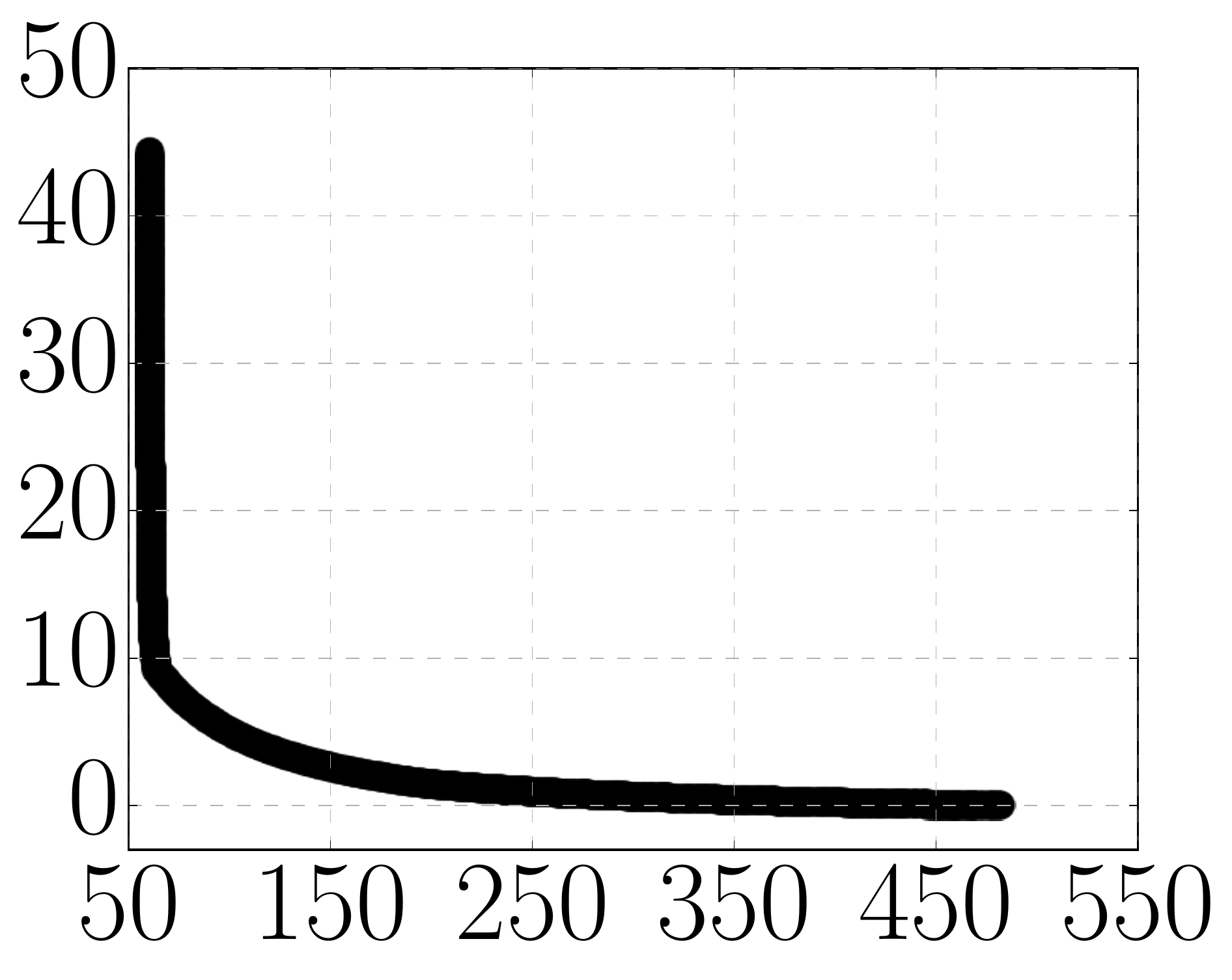}}
\subfloat[RE2-3-5]{\includegraphics[width=0.316\textwidth]{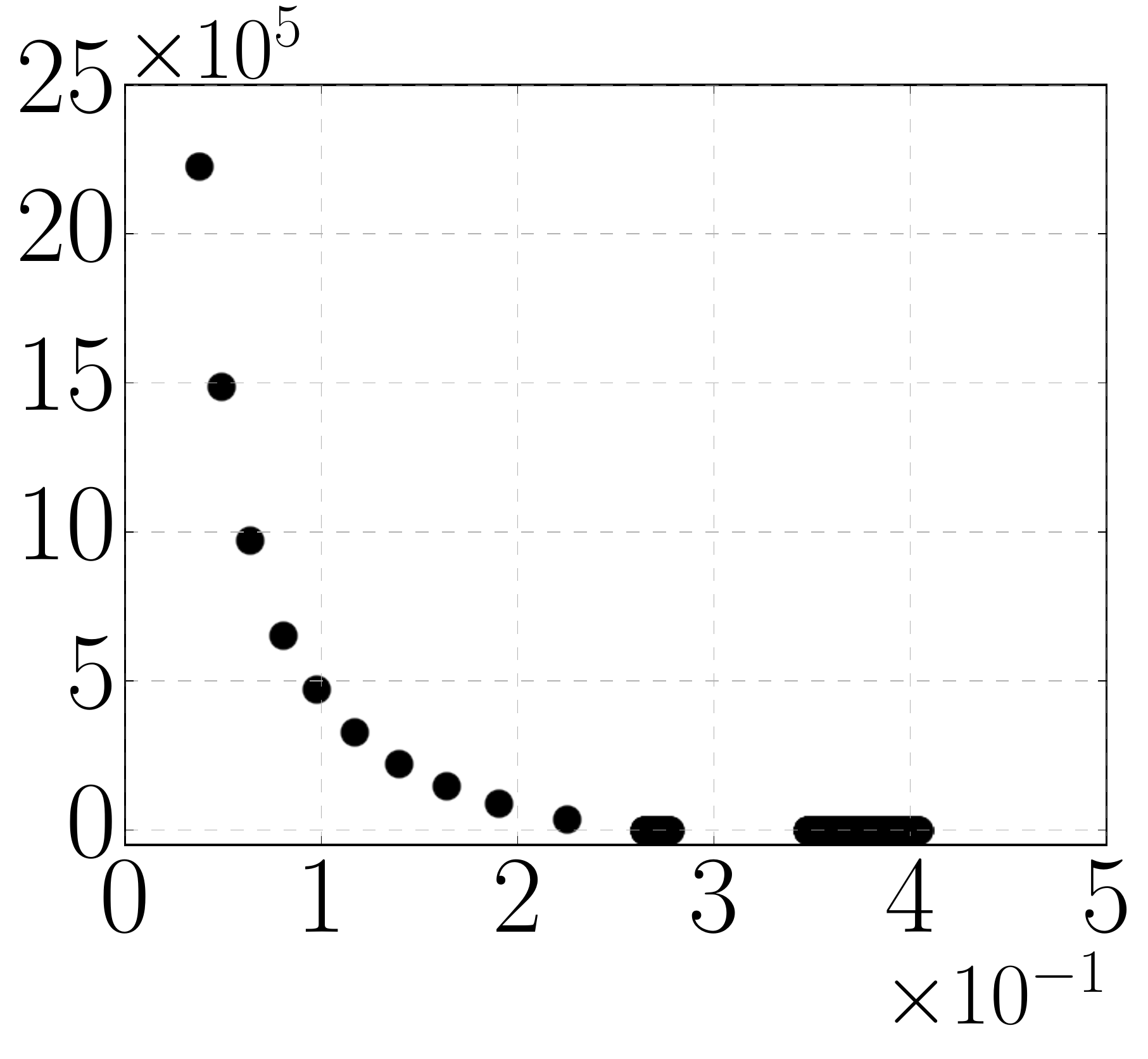}}
\caption{
\small
Approximated Pareto fronts of the RE2-4-1, ..., RE2-3-5 problems.
The x and y axis represent $f_1$ and $f_2$, respectively.
}
\label{fig:apf_2re}
\end{figure}

Figure \ref{fig:apf_2re} shows approximated Pareto fronts of the five two-objective RE problems (RE2-4-1, RE-2-3-2, RE2-4-3, RE2-2-4, and RE2-3-5).
As shown in Figure \ref{fig:apf_2re}, RE2-4-3 and RE2-3-5 have disconnected Pareto fronts whereas the other three problems have continuous Pareto fronts.
%
It should be noted that Figure \ref{fig:apf_2re} shows the approximated Pareto fronts, not the true Pareto fronts.
In contrast to synthetic test problems, it is almost impossible to obtain the true Pareto optimal solution set in a real-world problem.
While many synthetic test problems have regular shapes (except for some intentionally generated difficult problems \cite{LiZ09}), the RE problems in Figure \ref{fig:apf_2re} have somewhat irregular Pareto fronts.
This is because a synthetic problem is usually designed to have a typical Pareto front shape such as linear and convex.
However, real-world problems do not have such a nature.
The objective functions of most RE problems are also differently scaled.
For example, Figure \ref{fig:apf_2re} (a) shows that the first objective values in the approximated Pareto front are approximately in the range $[1.2 \times 10^3, 2.1 \times 10^3]$.
In contrast, the second objective values in the approximated Pareto front are approximately in the range $[2.8 \times 10^{-4}, 3.4 \times 10^{-3}]$.
Thus, the scale of the first and second objective values in RE2-4-1 is totally different.
We believe that most real-world problems have differently scaled objective values as in the RE problems.

The shape of the Pareto front of the RE2-4-1 problem is convex.
In the RE-2-3-2 problem, the two extreme regions of the Pareto front are locally concave and convex, respectively.
While the Pareto front shape of the RE2-4-3 problem seems to be convex at first glance, it is mixed and has disconnected regions (the top-left region close to the vertical axis is locally concave).
The RE2-2-4 problem has a convex Pareto front with a sharp knee point.
The RE2-3-5 problem has the mixed (linear and convex) and disconnected Pareto front.
The bottom-right region of the Pareto front is linear in Figure \ref{fig:apf_2re}(e).
Objective vectors on the disconnected region of the Pareto front are sparsely distributed.


\begin{figure}[htp]
\centering
\subfloat[RE3-3-1]{\includegraphics[width=0.49\textwidth]{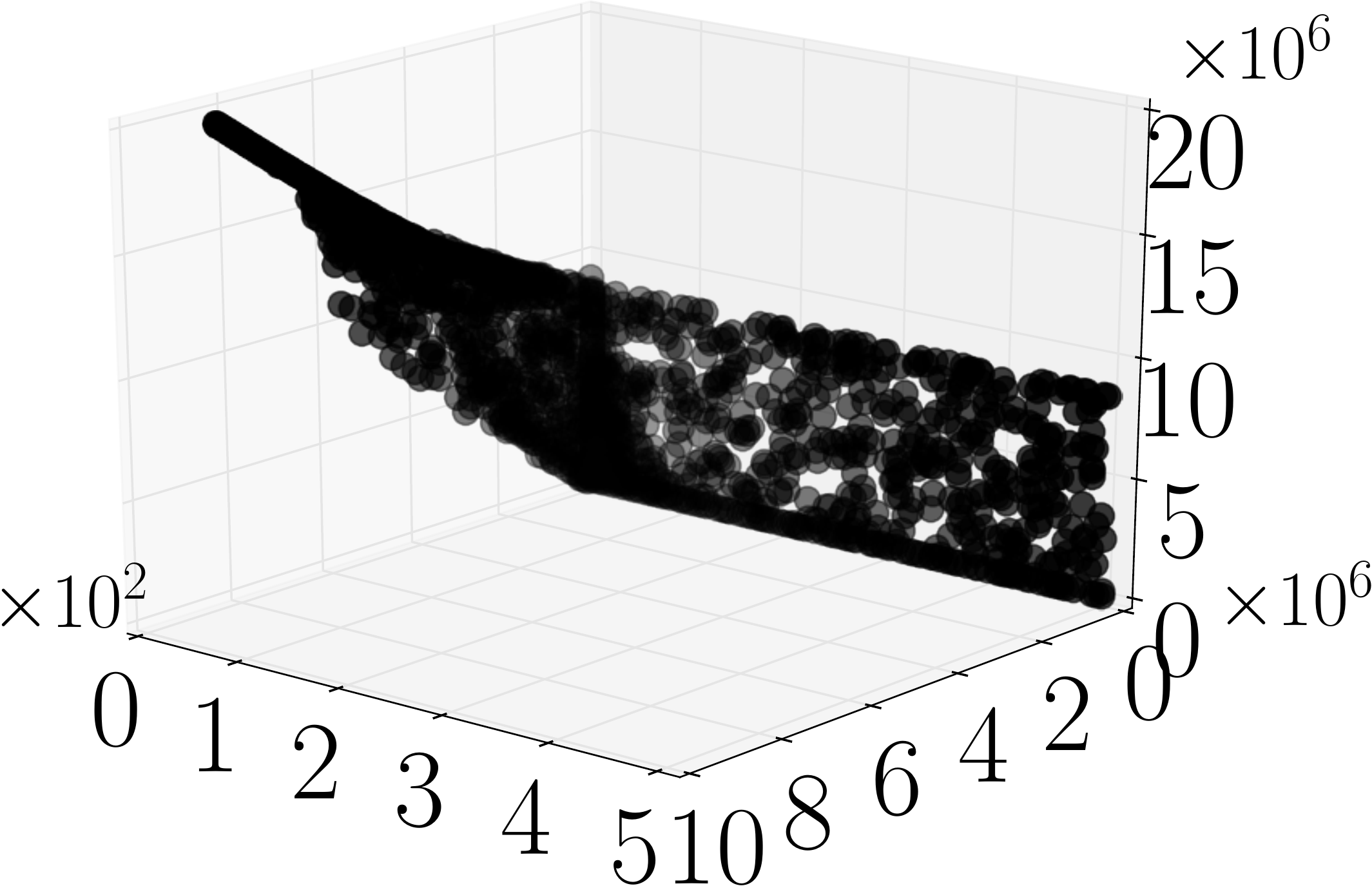}}
\subfloat[RE3-4-2]{\includegraphics[width=0.49\textwidth]{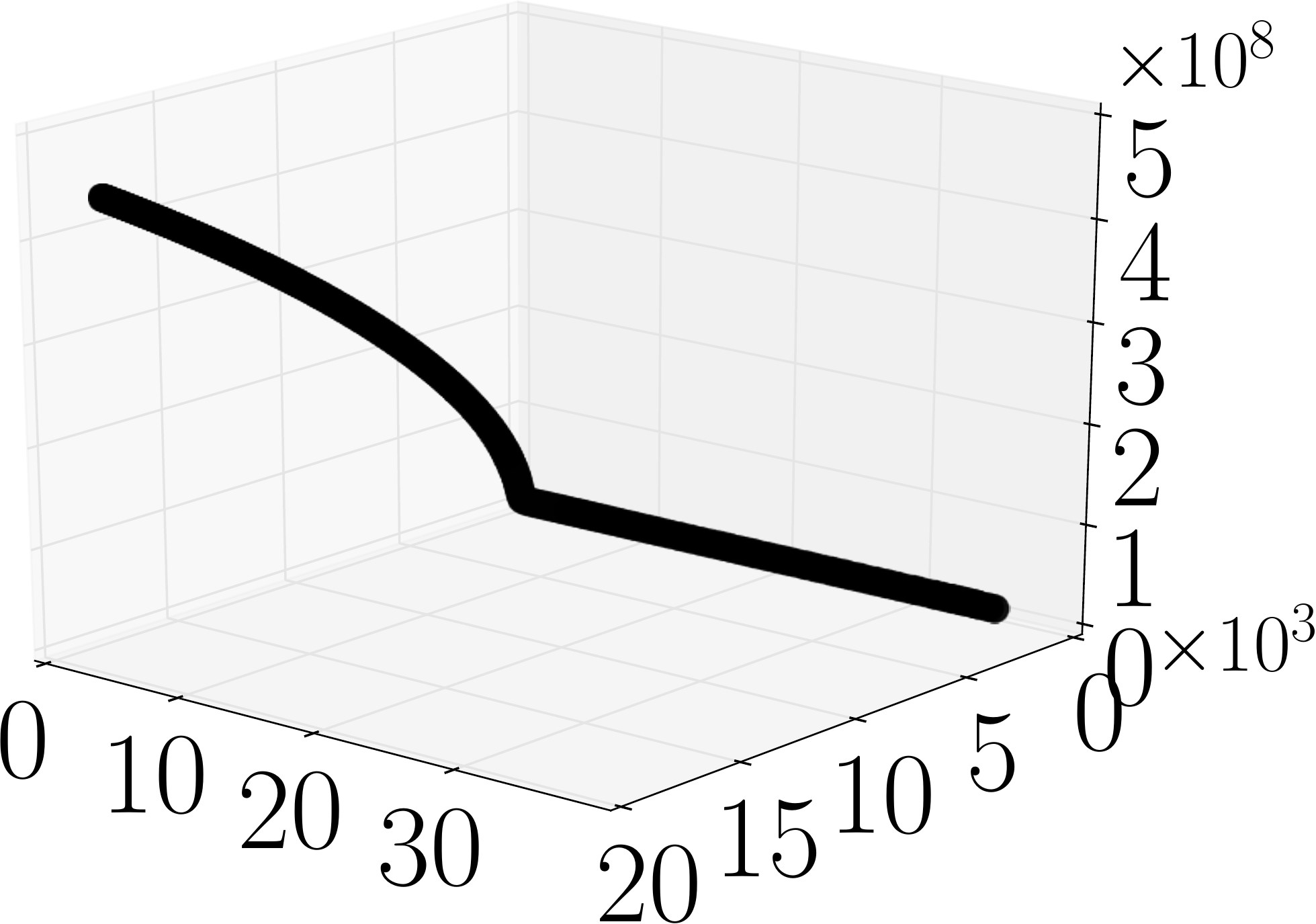}}
\\
\subfloat[RE3-4-3]{\includegraphics[width=0.49\textwidth]{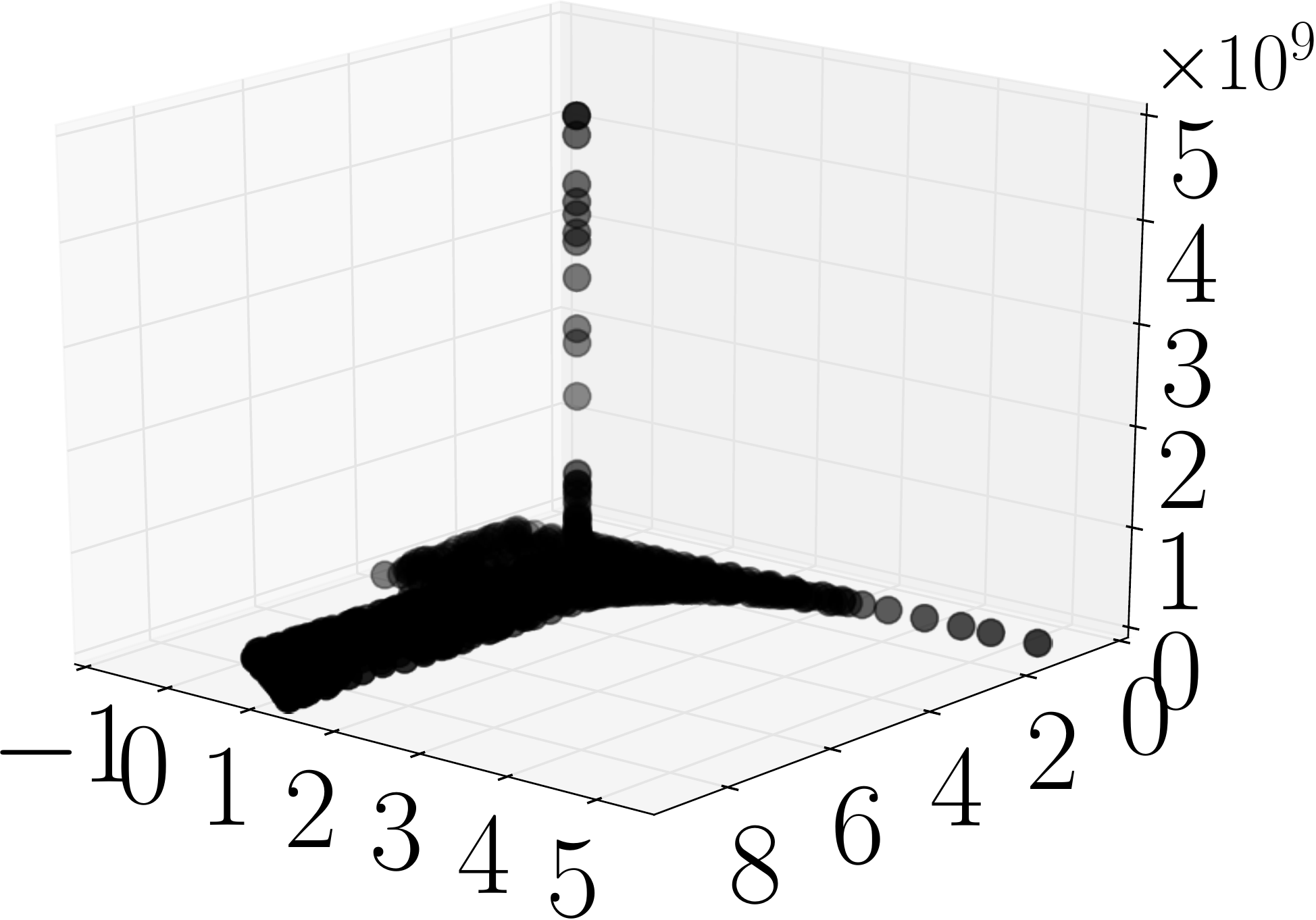}}
\subfloat[RE3-5-4]{\includegraphics[width=0.49\textwidth]{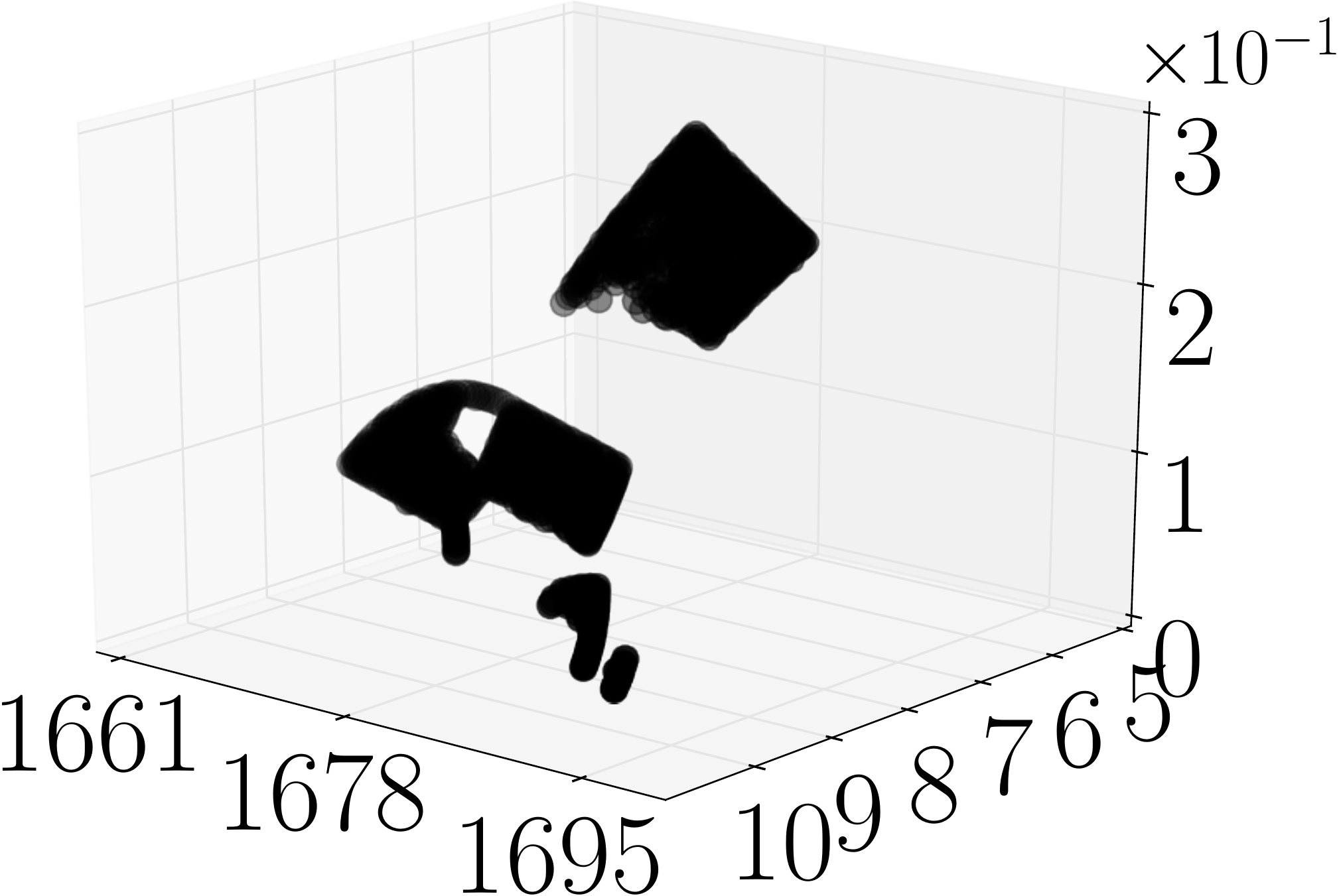}}
\\
\subfloat[RE3-4-6]{\includegraphics[width=0.49\textwidth]{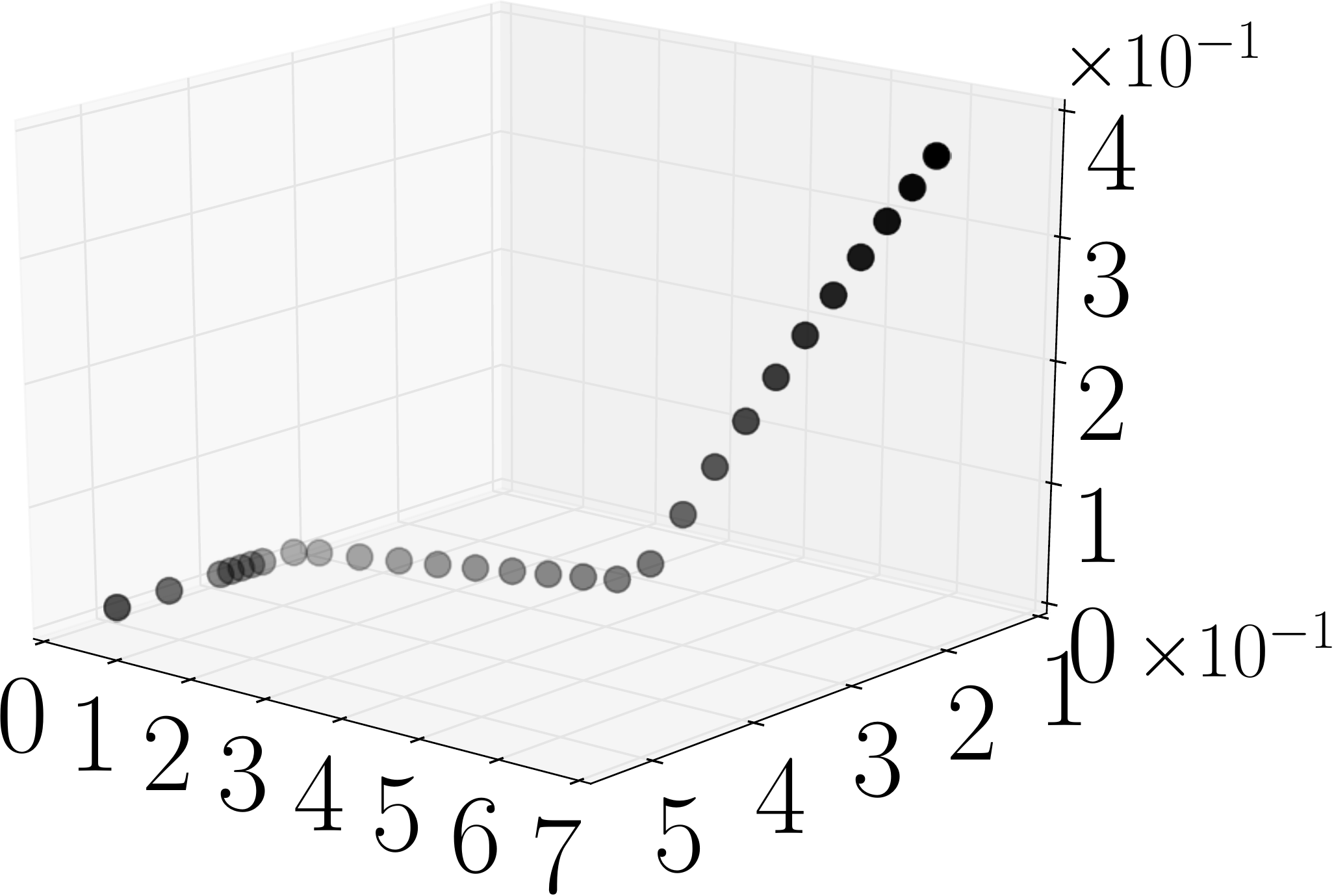}}
\subfloat[RE3-4-7]{\includegraphics[width=0.49\textwidth]{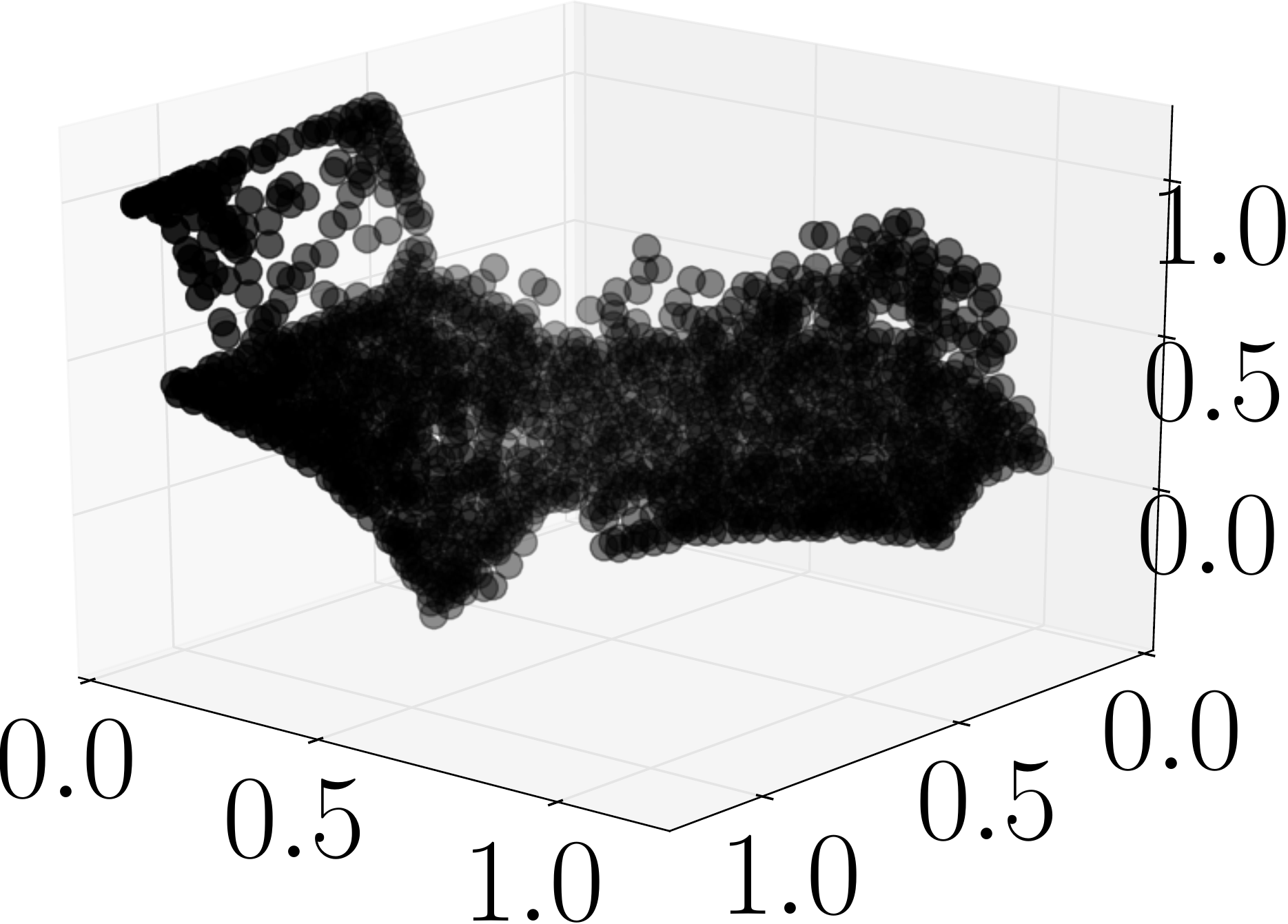}}
\caption{
\small
Approximated Pareto fronts of the RE3-3-1, ..., RE3-4-7 problems.
The $x$, $y$, and $z$ axis represent $f_1$, $f_2$, and $f_3$, respectively.
}
\label{fig:apf_3re}
\end{figure}




Fig. \ref{fig:apf_3re} shows approximated Pareto fronts of the six three-objective RE problems.
We do not show the approximated Pareto front of RE3-7-5, but it is similar to that of RE3-4-2.
While the approximated Pareto fronts of the three-objective RE problems can be still visualized by 3D graphics as  shown in Figure \ref{fig:apf_3re}, their shapes cannot be understood well.
Figs. S.1--S.11 in the supplementary file show the scatter matrix of the approximated Pareto front of RE3-3-1, RE3-4-2, RE3-4-3, RE3-5-4, RE3-7-5, RE3-4-6, RE3-4-7, RE4-7-1, RE4-6-2, RE6-3-1, and RE9-7-1, respectively.
Unfortunately, except for RE3-4-6, the scatter matrices do not provide clear shapes of the Pareto fronts.
This is because most RE problems have complicated Pareto fronts. 
It is usually difficult to clearly visualize Pareto front shapes of problems with four or more objectives \cite{TusarF15}.
Although parallel coordinates are frequently used to discuss the distribution of the objective vectors for problems with more than four objectives, they do not provide sufficient information about the shape of the Pareto front \cite{LiZY17}.


%

While recently proposed EMOAs perform well on problems with many objectives, they lack the ability to handle problems with irregular Pareto fronts \cite{IshibuchiSMN16,JiangY16}.
To address this issue, some EMOAs were designed to handle complex problems (e.g., \cite{JiangY16,YangJJ17,HuaJH18}).
The RE problems will be used as a proper problem suite to evaluate the performance of such EMOAs.

\section{Experimental settings}
\label{sec:experimental_settings}

We investigate the performance of representative EMOAs on the RE problem suite.
This section describes our experimental settings.
Results are reported in Section \ref{sec:experimental_results}.

\subsection{Performance indicators}
\label{sec:performance_indicators}


We use the hypervolume indicator (HV) \cite{ZitzlerT98} for the performance assessment.
HV is the only Pareto-compliant quality indicator known so far in the literature \cite{ZitzlerBT06}.
A large HV value indicates that a given solution set approximates the Pareto front well in terms of both convergence and diversity in the objective space.
In general, the Pareto optimal solutions and the Pareto front of a real-world problem are unavailable.
For this reason, we do not use quality indicators which require reference objective vectors (e.g., IGD \cite{CoelloS04} and the $\epsilon$-indicator  \cite{ZitzlerTLFF03}).
As in \cite{IshibuchiSMN16,YuanXWY16}, we set the $M$-dimensional reference vector for the HV calculation to $(1.1, ..., 1.1)^{\rm T}$ in the normalized objective space $[0,1]^M$.
Before calculating the HV value, we normalize the objective vector $\vector{f} (\vector{x})$ using the approximated ideal point $\vector{z}^{\rm ideal}$ and the approximated nadir point $\vector{z}^{\rm nadir}$ of each RE problem.
For each $i \in \{1, ..., M\}$, the $i$-th normalized objective value $f^{[0,1]}_i(\vector{x})$ of a solution $\vector{x}$ is obtained as follows: $f^{[0,1]}_i(\vector{x}) = (f_i(\vector{x}) - z^{\rm ideal}_i)/(z^{\rm nadir}_i - z^{\rm ideal}_i)$.




To obtain $\vector{z}^{\rm ideal}$ and $\vector{z}^{\rm nadir}$, we used L-SHADE \cite{TanabeF14CEC} and five EMOAs (NSGA-II \cite{DebAPM02}, MOEA/D-TCH \cite{ZhangL07}, MOEA/D-PBI \cite{ZhangL07}, IBEA \cite{ZitzlerK04}, and NSGA-III \cite{DebJ14}).
Only for these experiments, the number $n^{\rm max}$ of function evaluations was set to $100\,000$, and 31 independent runs were performed.
L-SHADE was the winner of the IEEE CEC2014 competition on single-objective real-parameter optimization.
We used the Java source code of L-SHADE provided by the authors of \cite{TanabeF14CEC}.
The default parameter setting was used for L-SHADE.
For each RE problem, we individually applied L-SHADE to each objective function.
For the five EMOAs, we used the same control parameter setting described in Section \ref{sec:settings_EMOAs}.
We applied the five EMOAs to each RE problem.
For L-SHADE, we maintained only the best-so-far solution for each run.
For the five EMOAs, we maintained only the final population for each run.

For each RE problem, let $\vector{A}$ be a set of non-dominated solutions obtained by the above-mentioned procedure.
We set the $i$-th objective value $z^{\rm ideal}_i$ of $\vector{z}^{\rm ideal}$ as follows: $z^{\rm ideal}_i = \min_{\vector{x} \in \vector{A}} f_i (\vector{x})$.
We also set the $i$-th objective value $z^{\rm nadir}_i$ of $\vector{z}^{\rm nadir}$ as follows: $z^{\rm nadir}_i = \max_{\vector{x} \in \vector{A}} f_i (\vector{x})$.

\subsection{Performance Score}
\label{sec:ps}

We used the performance score \cite{BaderZ11} to rank EMOAs on each RE problem.
Suppose that $n$ algorithms $A_1, ..., A_n$ are compared for a problem based on any performance indicator values obtained in multiple runs.
For each $i \in \{1, ..., n\}$ and $ j \in \{1, ..., n\} \backslash \{i\}$, let $\delta_{i,j} = 1$ if $A_j$ significantly outperforms $A_i$ based on the Wilcoxon rank-sum test with $p < 0.05$, otherwise $\delta_{i,j} = 0$.
The performance score $P(A_i)$ is defined as follows: $P(A_i) = \sum^{n}_{ j \in \{1, ..., n\} \backslash \{i\}} \delta_{i,j}$.
The score $P(A_i)$ represents the number of algorithms outperforming $A_i$. 
A small $P(A_i)$ value indicates that $A_i$ is not worse than many other algorithms. 


\subsection{EMOAs}
\label{sec:settings_EMOAs}

We examine the performance of NSGA-II \cite{DebAPM02}, MOEA/D \cite{ZhangL07}, IBEA \cite{ZitzlerK04}, SMS-EMOA \cite{BeumeNE07}, and NSGA-III \cite{DebJ14}, which are representative EMOAs based on dominance, decomposition, indicator, and reference vector, respectively.
We used the jMetal 4.5 framework \cite{DurilloN11} for algorithm implementation.
For NSGA-III, we used the source code implemented by the authors of \cite{YuanXWY16}.
As analyzed in \cite{IshibuchiISN16}, better performance of NSGA-III can be obtained by using the source code implemented by the authors of \cite{YuanXWY16} than the jMetal 5.0 source code \cite{NebroDV15}.
We conducted all experiments on a workstation with an Intel(R) Xeon(R) CPU E5-2620 v4@2.10GHz and 8GB RAM using the Ubuntu 16.04 OS.
The Java version that was used was OpenJDK version $1.8.0\_232$.
The number $n^{\rm max}$ of function evaluations was set to $10\,000$.
We performed 31 independent runs.



For all EMOAs, the population size $\mu$ was set to $100$, $105$, $120$, $210$, $182$, and $210$ for $M=2$, $3$, $4$, $6$, and $9$, respectively.
Since $\mu$ in NSGA-II must be a multiple of two, $\mu=106$ for $M=3$.
%
%
SBX crossover \cite{DebA95} and polynomial mutation \cite{DebA95} were used in all EMOAs.
Their control parameters were set as follows: $p_c = 1$, $\eta_c = 20$, $p_m = 1/D$, and $\eta_m = 20$.
A real-coded value was rounded to the nearest feasible value for discrete and integer variables in RE22, RE23, RE25, RE35, and RE36.
The additive epsilon indicator $I_{\epsilon+}$ \cite{ZitzlerK04} was used for IBEA.
According to \cite{ZhangL07}, the neighborhood size of MOEA/D was set to $20$.
MOEA/D decomposes $M$-objective problems into $N$ single-objective problems by using a set of weight vectors $\vector{W} = \{\vector{w}_1, ..., \vector{w}_N\}$ and a scalarizing function $g: \mathbb{R}^M \rightarrow \mathbb{R}$.
A set of weight/reference vectors of MOEA/D and NSGA-III were generated by the simplex-lattice design \cite{DasD98} for $M \leq 5$ and its two-layered version \cite{DebJ14} for $M \geq 6$.
We used the Tchebycheff function $g^{\rm tch}$ \cite{ZhangL07} and the PBI function $g^{\rm pbi}$ \cite{ZhangL07} in MOEA/D since they were used in many previous studies \cite{TrivediSSG17}.
The resulting algorithm versions are denoted as MOEA/D-TCH and MOEA/D-PBI, respectively.
The Tchebycheff function $g^{\rm tch}$ and the PBI function $g^{\rm pbi}$ are given as follows:
\begin{align}
\label{eqn:tchebycheff-mul}
g^{\rm tch}(\vector{x} | \vector{w}) &= \max_{i \in \{1, ..., M\}} \bigl\{ w_i |f_i (\vector{x}) - z^{\rm ideal}_i|  \bigr\},\\
\label{eqn:pbi}
g^{\rm pbi}(\vector{x} | \vector{w}) &= d_1 + \theta \, d_2,\\
\label{eqn:pbi_d1}
d_1 &= \frac{\| \left(\vector{f} (\vector{x}) - \vector{z}^{\rm ideal} \right)^{\rm T} \, \vector{w}\|}{\|\vector{w}\|},\\
\label{eqn:pbi_d2}
d_2 &= \left\| \vector{f} (\vector{x}) - \left(\vector{z}^{\rm ideal} +  d_1 \, \frac{\vector{w}}{\|\vector{w}\|} \right)\right\|,
\end{align}
where $\|\vector{a}\|$ is the Euclidean norm of a given vector $\vector{a}$.
The distance $d_1$ measures how well $\vector{f} (\vector{x})$ converges to the Pareto front.
The distance $d_2$ is the perpendicular distance between $\vector{f} (\vector{x})$ and $\vector{w}$.
Thus, $d_2$ measures how close $\vector{f} (\vector{x})$ and $\vector{w}$ are.
The penalty parameter $\theta$ strikes the balance between the convergence ($d_1$) and the diversity ($d_2$).
According to \cite{ZhangL07}, we set $\theta$ to $5$.
The simple normalization strategy described in \cite{ZhangL07} was introduced into MOEA/D-TCH and MOEA/D-PBI in order to handle the RE problems with totally different objective values.
As explained in Section \ref{sec:re_pf}, the objective values of most RE problems are differently scaled.
As demonstrated in \cite{DebJ14}, the use of the normalization strategy significantly improves the performance of EMOAs on problems with differently scaled objective values.

\section{Experimental results}
\label{sec:experimental_results}

\subsection{Performance comparison}
\label{sec:performance_comparison}

Table \ref{tab:results_re15} shows results of the six EMOAs on the 16 RE problems.
It should be noted that the rank of each EMOA in Table \ref{tab:results_re15} is based on the performance score value (see Section \ref{sec:ps}), not the raw HV value.
SMS-EMOA was applied only to the RE problems with two and three objectives due to its high computational cost.


Table \ref{tab:results_re15} indicates that there is no clear winner on all the two-objective RE problems.
IBEA, MOEA/D-TCH, and NSGA-II obtain the best HV value for RE2-4-1, RE2-4-3, and RE2-3-5, respectively.
SMS-EMOA performs the best on RE2-3-2 and RE2-2-4 in terms of HV.
%
Unlike the results of the two-objective RE problems, the best HV values are obtained by IBEA for 6 out of the 10 RE problems with three or more objectives.
As explained in Section \ref{sec:settings_EMOAs}, our IBEA uses the additive $\epsilon$ indicator $I_{\epsilon +}$, not the hypervolume difference indicator ($I_{HD}$) \cite{ZitzlerK04}.
Thus, our IBEA does not directly optimize HV.
NSGA-II and SMS-EMOA are competitive with each other on RE3-7-5 regarding the performance score with the Wilcoxon rank-sum test.
MOEA/D-PBI shows the worst performance on 10 out of the 16 RE problems.










\subsection{Discussion}
\label{sec:results_discussion}




\subsubsection{Inconsistency of the performance of NSGA-III and MOEA/D-PBI}
\label{sec:discussion_nsgaiii}

Table \ref{tab:results_re15} shows that NSGA-II outperforms NSGA-III on 10 out of the 16 RE problems in terms of the rank based on the performance score value.
While MOEA/D-PBI showed a good performance on many-objective problems in previous studies \cite{LiDZK15,DebJ14}, it does not work well in our study.

The reason is twofold.
First, NSGA-II and MOEA/D-PBI seemingly cannot handle irregular shape Pareto fronts well as described in Section \ref{sec:introduction}.
It was shown in \cite{JiangY16} that NSGA-III fails to find well-distributed solutions on some problems with convex or disconnected Pareto fronts.
As pointed out in \cite{IshibuchiDN16ppsn}, solutions obtained by MOEA/D-PBI with an intermediate $\theta$ value (e.g., $\theta=5$) are biasedly distributed on the center of the convex Pareto front.
For details of $\theta$, see Section \ref{sec:settings_EMOAs}.
The second reason is due to the setting of the maximum number of evaluations $n^{\rm max}$.
Previous studies \cite{TanabeIO17,TanabeIasoc18} show that some EMOAs (including NSGA-III and MOEA/D-PBI) do not work well for a small value of $n^{\rm max}$.
While $n^{\rm max}=10\,000$ in our study, it was set to a very large value in most recent studies \cite{TanabeIO17}.
For example, $n^{\rm max} = 550 \,000$ in the original paper of NSGA-III \cite{DebJ14}.
However, it is difficult to set $n^{\rm max}$ to a large value in some real-world problems.
This is because they require a computationally expensive simulation to evaluate a solution \cite{ObayashiSTH00}.
Since the RE problems are computationally cheap, $n^{\rm max}$ can be set to a large value (e.g., $n^{\rm max} = 550 \,000$).
However, it would be more practical to set $n^{\rm max}$ to a small value even on the RE problems so that the performance of EMOAs on computationally expensive problems can be estimated.

\begin{table}[t]
  \renewcommand{\arraystretch}{1}
\centering
\caption{\small 
  Results of the six EMOAs on the 16 RE problems. The table shows the mean HV values over 31 runs.
The numbers in parenthesis indicate the ranks of the six EMOAs based on the performance score value (see Section \ref{sec:ps}).
The first ranked results are represented by the bold font.
}
\label{tab:results_re15}
   {\scriptsize
\scalebox{0.88}[1]{
\begin{tabular}{ccccccccccc}
  \midrule
   & \raisebox{0.5em}{NSGA-II} & \shortstack{MOEA/D-\\TCH} & \shortstack{MOEA/D-\\PBI} & \raisebox{0.5em}{NSGA-III} & \raisebox{0.5em}{IBEA} & \shortstack{SMS-\\EMOA}\\
  \toprule

RE2-4-1 & 8.53e-01 (5) & 8.54e-01 (3) & 8.53e-01 (5) & 8.55e-01 (3) & {\ssmall \textbf{8.56e-01}} (1) & 8.55e-01 (2) \\
RE2-3-2 & 7.54e-01 (2) & 7.53e-01 (3) & 7.41e-01 (6) & 7.49e-01 (5) & 7.51e-01 (3) & {\ssmall \textbf{7.59e-01}} (1) \\
RE2-4-3 & 1.16e+00 (3) & {\ssmall \textbf{1.16e+00}} (1) & 1.12e+00 (6) & 1.16e+00 (4) & 1.16e+00 (2) & 1.15e+00 (5) \\
RE2-2-4 & 1.17e+00 (2) & 1.17e+00 (3) & 1.09e+00 (6) & 1.17e+00 (3) & 1.16e+00 (5) & {\ssmall \textbf{1.17e+00}} (1) \\
RE2-3-5 & {\ssmall \textbf{1.08e+00}} (1) & 1.08e+00 (3) & 1.04e+00 (6) & 1.08e+00 (5) & 1.08e+00 (3) & 1.08e+00 (2) \\
RE3-3-1 & 1.33e+00 (3) & {\ssmall \textbf{1.33e+00}} (1) & 1.32e+00 (6) & 1.33e+00 (2) & 1.33e+00 (4) & 1.33e+00 (5) \\
RE3-4-2 & {\ssmall \textbf{1.33e+00}} (1) & 1.31e+00 (4) & 1.28e+00 (6) & 1.33e+00 (2) & 1.33e+00 (3) & 1.30e+00 (5) \\
RE3-4-3 & 9.05e-01 (5) & 9.55e-01 (3) & 8.81e-01 (5) & 9.78e-01 (2) & 9.28e-01 (4) & {\ssmall \textbf{1.00e+00}} (1) \\
RE3-5-4 & 1.03e+00 (3) & 1.00e+00 (5) & 9.60e-01 (6) & 1.02e+00 (4) & {\ssmall \textbf{1.04e+00}} (1) & {\ssmall \textbf{1.04e+00}} (1) \\
RE3-7-5 & {\ssmall \textbf{1.30e+00}} (1) & 1.30e+00 (3) & 1.28e+00 (6) & 1.29e+00 (4) & 1.30e+00 (5) & {\ssmall \textbf{1.30e+00}} (1) \\
RE3-4-6 & {\ssmall \textbf{1.01e+00}} (1) & 9.94e-01 (5) & 9.92e-01 (5) & 1.00e+00 (4) & 1.00e+00 (2) & 1.00e+00 (2) \\
RE3-4-7 & 7.89e-01 (5) & 8.07e-01 (2) & 8.02e-01 (3) & 7.92e-01 (4) & {\ssmall \textbf{8.14e-01}} (1) & 7.89e-01 (5) \\
RE4-7-1 & 6.79e-01 (5) & 7.19e-01 (2) & 7.20e-01 (2) & 7.16e-01 (4) & {\ssmall \textbf{7.64e-01}} (1) & - \\
RE4-6-2 & 7.49e-01 (3) & 7.66e-01 (2) & 7.33e-01 (3) & 7.41e-01 (5) & {\ssmall \textbf{7.86e-01}} (1) & - \\
RE6-3-1 & 1.05e+00 (4) & 1.09e+00 (3) & 9.62e-01 (5) & 1.12e+00 (2) & {\ssmall \textbf{1.18e+00}} (1) & - \\
RE9-7-1 & 4.39e-02 (3) & 5.58e-02 (2) & 2.11e-02 (5) & 4.17e-02 (4) & {\ssmall \textbf{6.61e-02}} (1) & - \\
  \midrule
\end{tabular}
}
 }
\end{table}

\section{Conclusion}
\label{sec:conclusion}


While real-world problems are important in evaluating the performance of EMOAs, they are difficult to use due to some obstacles.
To fill this gap, we presented the RE problem suite, which consists of 16 real-world multi-objective optimization problems.
We provided Java, C, and Matlab source codes of the RE problem suite as the supplementary materials so that researchers can use them as standard benchmark problems.
We also presented the comparison of six EMOAs on the RE problems.

\begin{figure}[t]
\centering
\includegraphics[width=0.5\textwidth]{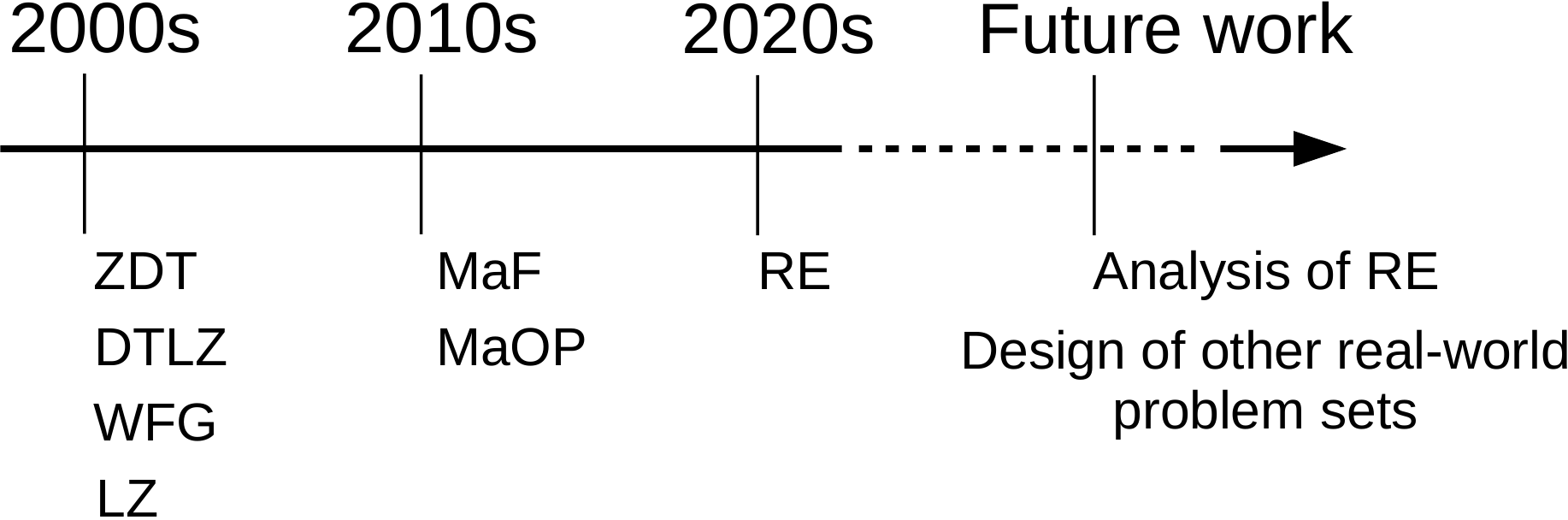}
\caption{
\small
Progress on multi-objective continuous test problem sets.
}
\label{fig:map}
\end{figure}

Figure \ref{fig:map} shows the progress on multi-objective continuous test problem sets.
For details, refer to \cite{HubandHBW06,ZapotecasCAT18}.
In the 2000s, various test problem sets have been proposed in the literature such as ZDT \cite{ZitzlerDT00}, DTLZ \cite{DebTLZ05}, WFG \cite{HubandHBW06}, and LZ \cite{LiZ09} problem suites.
In the 2010s, more sophisticated problem sets have been proposed, including MaF \cite{ChengLTZYJY17} and MaOP \cite{LiDZSC19}.
However, most synthetic test problems have some undesirable features as discussed in Section \ref{sec:introduction}.
We believe that the RE problem set can address these issues.
Future research topics include the performance evaluation of other state-of-the-art EMOAs and further analysis of each RE problem (e.g., shape of the feasible region of each problem in the objective space, distance from initial solutions to the Pareto front, difficulty of convergence and diversification).
Designing other benchmark sets (e.g., a set of dynamic multi-objective real-world problems) is also an avenue for future work.
We also hope that the RE problem set is helpful in developing efficient EMOAs for a general computational intelligence aided design framework in the smart design process.

 \section*{Acknowledgments}


 This work was supported by National Natural Science Foundation of China (Grant No. 61876075), the Program for Guangdong Introducing Innovative and Enterpreneurial Teams (Grant No. 2017ZT07X386), Shenzhen Peacock Plan (Grant No. KQTD2016112514355531), the Science and Technology Innovation Committee Foundation of Shenzhen (Grant No. ZDSYS201703031748284), the Program for University Key Laboratory of Guangdong Province (Grant No. 2017KSYS008).
 


 

\section*{References}

\bibliography{reference}

\end{document}